\pdfoutput=1
\documentclass[pdflatex,sn-mathphy]{sn-jnl}

\usepackage{natbib}
\usepackage{geometry}
\usepackage{placeins}
\usepackage{graphicx}
\usepackage{caption}
\usepackage{multirow}%
\usepackage{amsmath,amssymb,amsfonts}%
\usepackage{amsthm}%
\usepackage{mathrsfs}%
\usepackage[title]{appendix}%
\usepackage{xcolor}%
\usepackage{textcomp}%
\usepackage{manyfoot}%
\usepackage{booktabs}%
\usepackage{algorithm}%
\usepackage{algorithmicx}%
\usepackage{algpseudocode}%
\usepackage{listings}%
\usepackage{pgfplots}
\pgfplotsset{compat=1.18}

\theoremstyle{thmstyleone}%

\theoremstyle{thmstyletwo}%

\theoremstyle{thmstylethree}%

\begin{document}
\pagenumbering{gobble}

\title[Fraud Detection System for Banking Transactions]{Fraud Detection System for Banking Transactions}

\author[1]{\fnm{Ranya} \sur{Batsyas}}\email{ranyabatsyas@gmail.com}

\author[2]{\fnm{Ritesh} \sur{Yaduwanshi}}\email{riteshyaduwanshi@igdtuw.ac.in}

\affil[1]{\orgdiv{Department of AI DS}, \orgname{IGDTUW}, \orgaddress{\city{Delhi}, \country{India}}}

\affil[2]{\orgdiv{Department of AI DS}, \orgname{IGDTUW}, \orgaddress{\city{Delhi}, \country{India}}}

\abstract{The expansion of digital payment systems has heightened both the scale and intricacy of online financial transactions, thereby increasing vulnerability to fraudulent activities. Detecting fraud effectively is complicated by the changing nature of attack strategies and the significant disparity between genuine and fraudulent transactions. This research introduces a machine learning-based fraud detection framework utilizing the PaySim synthetic financial transaction dataset. Following the CRISP-DM methodology, the study includes hypothesis-driven exploratory analysis, feature refinement, and a comparative assessment of baseline models such as Logistic Regression and tree-based classifiers like Random Forest, XGBoost, and Decision Tree. To tackle class imbalance, SMOTE is employed, and model performance is enhanced through hyperparameter tuning with GridSearchCV. The proposed framework provides a robust and scalable solution to enhance fraud prevention capabilities in FinTech transaction systems.}


\maketitle

\section{Introduction}\label{sec:intro}
\subsection{Background}

The rapid proliferation of digital payment systems and e-commerce platforms has significantly transformed the global financial landscape, facilitating millions of transactions daily while concurrently heightening susceptibility to financial fraud. As transaction volumes increase, fraudulent activities have become more prevalent and sophisticated, posing substantial risks to consumers, merchants, and FinTech companies. In 2023, global losses from online payment and e-commerce fraud were estimated  at US\$48 billion with payment-card fraud alone accounting for about US\$33.8 billion. It is further estimated that nearly 3--4\% of all digital transactions worldwide may be fraudulent, underscoring the magnitude of the issue. Traditional rule-based fraud detection systems struggle to adapt to evolving fraud strategies and are further constrained by the highly imbalanced nature of financial transaction data, where fraudulent cases are infrequent. This imbalance often results in suboptimal detection performance and elevated false-positive rates. 

To address these challenges, this project proposes a machine learning–based fraud detection framework utilizing the PaySim synthetic transaction dataset and adhering to the  CRISP-DM (Cross-Industry Standard Process for Data Mining) methodology . Multiple classification models—including Logistic Regression, Decision Tree, Random Forest, and XGBoost—are evaluated, with SMOTE applied to enhance minority-class representation. Model performance is primarily assessed using the F1 score, and the best-performing model is further optimized using GridSearchCV to develop a robust and scalable fraud detection system suitable for FinTech environments.

\subsection{Literature Review }

The detection of financial fraud has evolved from static, rule-based systems to data-driven machine learning methodologies. This shift addresses the increasing complexity, adaptability, and data imbalance characteristic of fraudulent activities, with behavioral modeling playing a pivotal role in contemporary approaches. \citet{kumar2022}  illustrate that integrating behavioral features—such as transaction frequency, timing, and spending deviations—greatly enhances detection accuracy by identifying subtle anomalies in user behavior. 

Major challenges in fraud detection, such as class imbalance, feature sparsity, and concept drift, stressing the need for preprocessing strategies like resampling, cost-sensitive learning, and adaptive model updates to ensure robustness in operational settings. A significant body of research identifies ensemble learning as a leading method for fraud detection in imbalanced datasets. Further emphasizing the importance of temporal and relational data,\citet{cochrane2021}  analyze transaction streams to extract structured and hypothesis-driven features. Their research underscores the added value of contextual information—for example, sequences of transactions and relational links between entities—that help anticipate fraudulent actions earlier than conventional methods.

\citet{fan2025} and \citet{kumar2020} demonstrate that tree-based ensemble models, including Random Forests and XGBoost, consistently outperform linear and kernel-based classifiers, especially when used with SMOTE-based oversampling techniques. These findings are supported by \citet{bhattacharyya2011} , whose comparative study underscores the superior predictive capability of ensemble methods for credit card fraud detection. \citet{makki2019}  and \citet{achary2023}  further confirm that imbalance-aware learning strategies improve minority class recognition while reducing false positives. Boosting and cost-sensitive frameworks have gained traction due to their ability to explicitly prioritize fraudulent instances during training. \citet{Bala2024}  show the effectiveness of XGBoost for online fraud detection, while \citet{randhawa2018}  propose cost-sensitive boosting algorithms that adaptively reweight misclassification costs to enhance fraud recall. These approaches build on the foundational work of \citet{sun2007}, who established cost-sensitive boosting as a principled solution for imbalanced classification problems. \citet{varmedja2019}  add to these insights with their comprehensive survey that categorizes ML approaches into supervised, unsupervised, and hybrid methods, emphasizing the ongoing relevance of SMOTE and ensemble classifiers, particularly random forests. Recent research extends fraud detection beyond static tabular modeling by incorporating temporal and relational dependencies. 

\citet{zhang2025}  introduce graph neural network-based methods that capture complex user interactions and fraud communities, enabling the detection of coordinated and organized fraud schemes. 

\citet{maniraj2019credit}  focus on ensemble designs aimed at maximizing fraud recall, while \citet{V2012}  demonstrate that classical classifiers such as SVMs remain effective when enhanced with behavior-driven feature engineering. 

From 2024 onward, the literature increasingly emphasizes deep learning and hybrid architectures to address temporal dynamics and concept drift. \citet{feng2025hybrid}  proposes a BiLSTM–Transformer framework that significantly improves fraud recall in mobile banking environments. \citet{btoush2025}  show that stacking deep feature extractors with gradient boosting and explainability mechanisms enhances minority class precision on highly imbalanced datasets. Similarly, 
\citet{tsai2025} , \citet{gr2024}, and \citet{khosravi2023} demonstrate that attention-enhanced and sequence-aware models outperform traditional ML approaches in detecting low-frequency and evolving fraud patterns.

\section{Methodology}\label{sec:methodology}

This study employs the CRISP-DM framework to analyze and model fraudulent financial transactions through its six stages: Business Understanding, Data Understanding, Data Preparation, Modeling, Evaluation, and Deployment. This structured methodology ensures alignment with industry best practices. The primary aim is to develop a robust fraud detection system that achieves high recall with acceptable precision, learns behavioral patterns of fraudulent activity, addresses class imbalance and noisy data, and provides interpretable insights to support risk analysis.
\begin{figure}[h]
    \centering
    \includegraphics[scale=0.8]{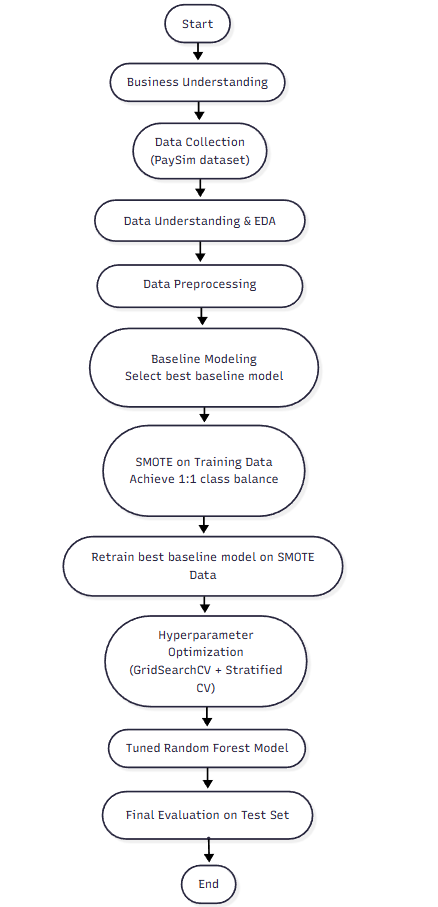}
    \caption{Methodology Flowchart}
\end{figure}

\subsection{Data Collection}\label{subsec:datacollection}

We have used the Kaggle dataset ``Synthetic Financial Datasets For Fraud Detection'' generated by the PaySim mobile money simulator. The dataset includes a wide range of features such as transaction amount, transaction type, time of transaction, sender and receiver's bank account details, etc. This dataset has 10 parameters and nearly 600,000 transactions.

The outcome variable is a binary label indicating whether a transaction is fraudulent (1) or legitimate (0). The primary challenge is the significant class imbalance: only 8,213 fraudulent cases (0.129\%) exist among 6,362,620 legitimate transactions.

\subsection{Data Understanding}\label{subsec:featureeng}

\subsubsection{Hypothesis Creation}

Domain specific hypotheses guide the exploratory analysis:

\begin{itemize}
\item \textbf{H1 - High-Value Transactions:} The transaction amount is positively correlated with the likelihood of fraud. Fraudsters exploit high-value transfers to maximize the monetary gain before detection.

\item \textbf{H2 - Transaction Type Risk:} Fraud is concentrated in TRANSFER and CASHOUT transaction types. These types allow for quick fund movement and conversion to cash, facilitating illegal liquidity extraction.

\item \textbf{H3 - Mule Account Networks:} Recipient accounts that frequently appear in fraudulent transactions suggest ``mule'' accounts---intermediaries for channeling illicit funds through fraud networks.
\end{itemize}

\subsubsection{Extrapolatory Data Analysis}

Univariate analysis indicated that fraudulent activities were solely found in the TRANSFER 49.86\% and CASHOUT (50.14\%) categories, with no instances of fraud in other transaction types, thus supporting Hypothesis 2 and suggesting a sequential fraud pattern where funds are transferred and then withdrawn.

\begin{figure}[h]
    \centering
    \includegraphics[scale=0.5]{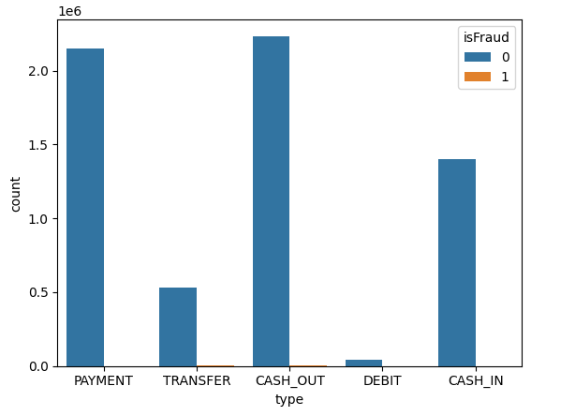}
    \caption{Bar plot showing transaction counts by type, with fraud appearing only in TRANSFER and CASHOUT.}
\end{figure}
\FloatBarrier

\begin{figure}[h]
    \centering
    \includegraphics[scale=0.3]{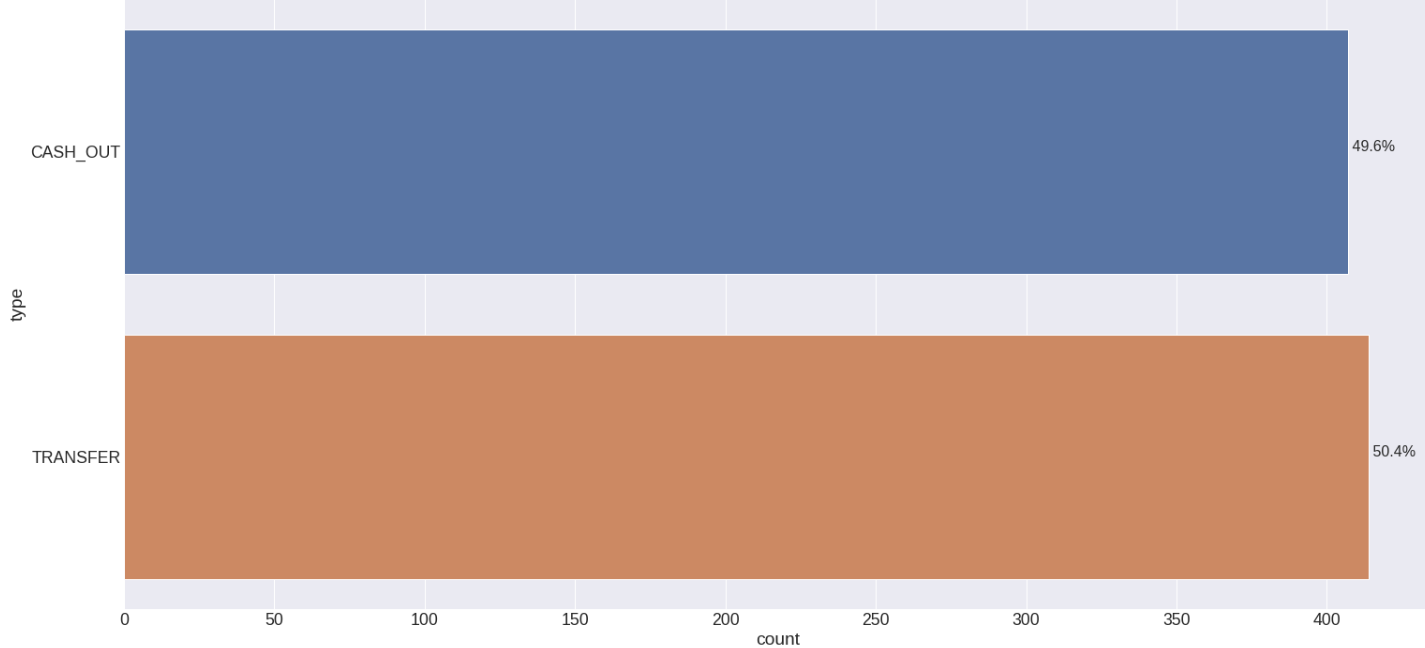}
    \caption{Horizontal bar plot showing the nearly equal split of all fraud cases between TRANSFER (49.86\%) and CASHOUT (50.14\%) transactions.}
\end{figure}
\FloatBarrier

The analysis of class distribution highlighted a significant imbalance, with fraudulent transactions making up just 0.129\% of the total (a ratio of 1:775). 
\begin{figure}[h]
    \centering
    \includegraphics[scale=0.3]{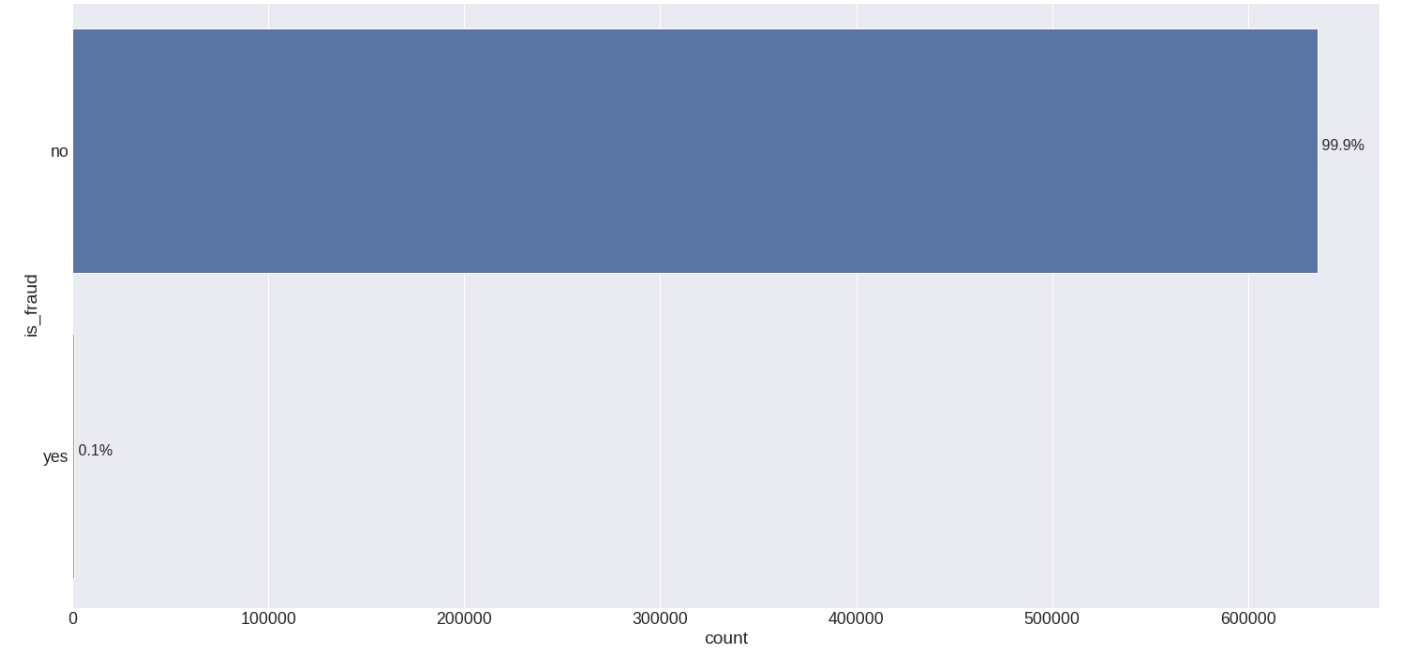}
    \caption{Horizontal bar plot demonstrating extreme class imbalance.}
\end{figure}
\FloatBarrier

Additionally, correlation analysis revealed a positive link between the amount of the transaction and the likelihood of fraud, which supports Hypothesis 1.
\begin{figure}[h]
    \centering
    \includegraphics[scale=0.5]{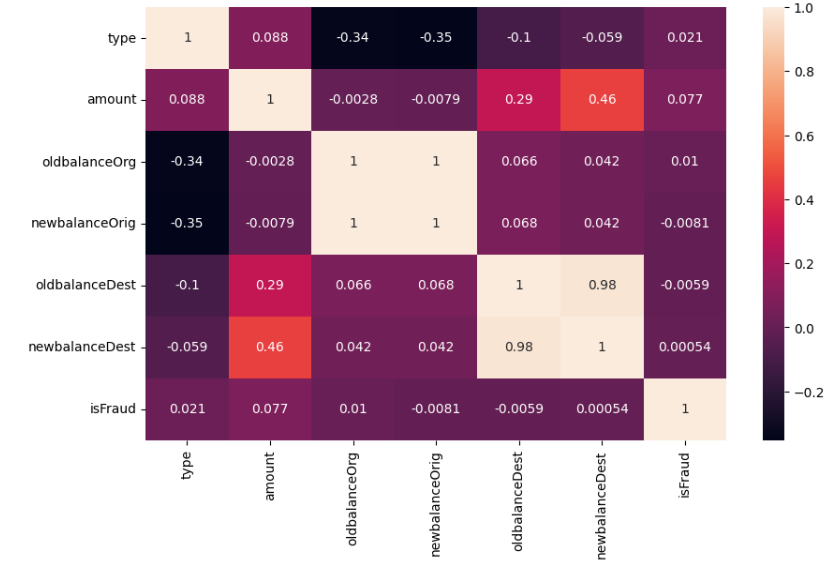}
    \caption{Correlation matrix demonstrating relationships between variables.}
\end{figure}
\FloatBarrier

\subsection{Data Preparation}\label{subsec:dataprep}

\subsubsection{Feature Engineering}

After evaluating their empirical variance and predictive value, four features were removed: step, nameOrig, nameDest, and isFlaggedFraud. 
In the PaySim dataset, the step feature acts as a monotonically increasing temporal index that systematically cycles through all transactions, showing no significant separation between classes. Fraudulent and non-fraudulent transactions are almost evenly distributed across its range, with nearly identical class-wise means and standard deviations. 

Similarly, isFlaggedFraud remains nearly constant ($\approx$ 0\ for most records), leading to almost zero variance and offering no additional information beyond the target label, making it redundant and potentially problematic due to multicollinearity and noise. 

On the other hand, nameOrig and nameDest are raw account identifiers that would need aggregation or graph-based transformations to be informative, as they do not inherently encode stable behavioral patterns. 

As a result, the final feature set retained six predictors: type (categorical transaction category), amount (numeric transaction value), oldbalanceOrg and newbalanceOrig (sender’s balances before and after the transaction), and oldbalanceDest and newbalanceDest (recipient’s balances before and after the transaction).

\subsubsection{Categorical Encoding}

Categorical encoding was applied using scikit-learn's LabelEncoder, converting transaction types into integer values (CASH-IN \(\rightarrow\) 0, CASH-OUT \(\rightarrow\) 1, DEBIT \(\rightarrow\) 2, PAYMENT \(\rightarrow\) 3, TRANSFER \(\rightarrow\) 4).

\subsection{Predictive Modelling}\label{subsec:modeling}

\subsubsection{Train-Test Split}

The dataset was then divided using stratified random sampling with a 90:10 train-test split. This stratification produced an X\_train of shape (5,726,358, 6) with Y\_train having 5,718,966 non-fraud (0) and 7,392 fraud (1) samples, and an X\_test of shape (636,262, 6) with Y\_test containing 635,441 non-fraud (0) and 821 fraud (1) samples,  preserving the original fraud prevalence rate of 0.12\%. 

\subsubsection{Model Development and Training}

To establish performance benchmarks and identify the best candidate for hyperparameter tuning, four baseline classification algorithms were assessed on the imbalanced training dataset:

\begin{itemize}
\item \textbf{Logistic Regression} was applied using L-BFGS optimization (max\_iter=1000) to create a linear baseline and define interpretable decision boundaries.

\item \textbf{Decision Tree Classifier} (max\_depth=None, random\_state=42) was used to capture non-linear feature interactions and establish interpretable decision rules.

\item \textbf{Random Forest} (n\_estimators=15, random\_state=42) was evaluated to utilize ensemble bagging for variance reduction and better generalization.

\item \textbf{XGBoost} (max\_depth=3, learning\_rate=0.1, random\_state=42) was configured with native class weighting capability via scale\_pos\_weight to address the imbalance.
\end{itemize}

\subsubsection{SMOTE (Synthetic Minority Oversampling Technique)}

In order to tackle the significant class imbalance, SMOTE was applied solely to the training dataset. This technique generated synthetic samples of the minority class through k-nearest neighbors interpolation, effectively altering the original imbalance ratio of about 755:1 to a balanced 1:1. The Random Forest model was then trained on this SMOTE-adjusted data, utilizing 15 estimators, parallel processing, and a fixed random seed to ensure the results could be replicated. 

\subsubsection{Hyperparameter Optimization (HPO)}

The Random Forest classifier underwent hyperparameter tuning via grid search, employing stratified 3-fold cross-validation on the SMOTE-adjusted training data. The search space encompassed crucial parameters such as : n\_estimators (50, 100), max\_depth (None, 10), min\_samples\_split (2), min\_samples\_leaf (1, 2), max\_features (`sqrt'), and class\_weight (`balanced'), resulting in 16 potential configurations. 

Class\_weight='balanced' specifically modified the penalty for misclassification costs to ensure that errors involving fraud cases are penalized more severely than those involving legitimate cases, even after SMOTE oversampling.

The F1-score served as the optimization metric to balance precision and recall in the context of class imbalance. Stratification maintained class distribution across the folds, and the optimal model utilized class-weighted learning to enhance the SMOTE-based resampling before the final evaluation.

\section{Results}\label{sec:results}

\subsection{Baseline Models}\label{subsec:baseline}

\subsubsection{Logistic Regression}

Logistic Regression achieved an accuracy of 99.91\%; however, the model's ability to detect fraud was limited, with a recall of just 0.43\% and an F1-score of 0.57, indicating a significant bias towards the majority class despite the high overall accuracy.

\begin{table}[h]
\caption{Logistic regression classification report}\label{tab:logistic}
\begin{tabular}{@{}lcccc@{}}
\toprule
& \textbf{Precision} & \textbf{Recall} & \textbf{F1-score} & \textbf{Support} \\
\midrule
Non-Fraud [0] & 1.00 & 1.00 & 1.00 & 635441 \\
Fraud [1] & 0.85 & 0.43 & 0.57 & 821 \\
\midrule
Accuracy & & & 1.00 & 636262 \\
Macro avg & 0.92 & 0.72 & 0.79 & 636262 \\
Weighted avg & 1.00 & 1.00 & 1.00 & 636262 \\
\botrule
\end{tabular}
\footnotetext{\textbf{Logistic regression model:} [[635377, 64], [467, 354]]}
\footnotetext{\textbf{Accuracy score on Test Data:} 0.9991654381371197\%}
\end{table}
\FloatBarrier

\subsubsection{Decision Tree Classifier}

Decision Tree Classifier (max\_depth=None, random\_state=42) showed a marked improvement with a 97\% fraud recall and an F1-score of 0.84, though the unrestricted tree depth raises concerns about overfitting on the large training set.

\begin{table}[h]
\centering
\caption{Decision tree classification report}\label{tab:decision_tree}
\begin{tabular}{lcccc}
\toprule
& \textbf{Precision} & \textbf{Recall} & \textbf{F1-score} & \textbf{Support} \\
\midrule
Non-Fraud [0] & 1.00 & 1.00 & 1.00 & 635441 \\
Fraud [1] & 0.74 & 0.97 & 0.84 & 821 \\
\midrule
Accuracy & & & 1.00 & 636262 \\
Macro avg & 0.87 & 0.99 & 0.92 & 636262 \\
Weighted avg & 1.00 & 1.00 & 1.00 & 636262 \\
\botrule
\end{tabular}
\footnotetext{\textbf{Decision Tree Classifier Model:} [[635153, 288], [22, 799]]}
\footnotetext{\textbf{Accuracy score on Test Data:} 0.9995127793267553\%}
\end{table}
\FloatBarrier

\subsubsection{Random Forest}

Random Forest (n\_estimators=15, random\_state=42) achieved the highest F1-score of 0.87, along with outstanding fraud precision (97\%) and a 79\% recall, indicating an effective balance between precision and recall for the severe class imbalance issue.

\begin{table}[h]
\centering
\caption{Random forest classification report}\label{tab:random_forest}
\begin{tabular}{lcccc}
\toprule
& \textbf{Precision} & \textbf{Recall} & \textbf{F1-score} & \textbf{Support} \\
\midrule
Non-Fraud [0] & 1.00 & 1.00 & 1.00 & 635441 \\
Fraud [1] & 0.97 & 0.79 & 0.87 & 821 \\
\midrule
Accuracy & & & 1.00 & 636262 \\
Macro avg & 0.98 & 0.90 & 0.94 & 636262 \\
Weighted avg & 1.00 & 1.00 & 1.00 & 636262 \\
\botrule
\end{tabular}
\footnotetext{\textbf{Random forest model:} [[635421, 20], [169, 652]]}
\footnotetext{\textbf{Accuracy score on Test Data:} 0.9997029525572798\%}
\end{table}
\FloatBarrier

\subsubsection{XGBoost}

XGBoost (max\_depth=3, learning\_rate=0.1, random\_state=42) achieved a competitive ROC-AUC (0.96) and precision (98\%); however, XGBoost showed lower recall (70\%) and an F1-score of 0.82 compared to Random Forest.
\begin{table}[h]
\centering
\caption{XGBoost classification report}\label{tab:xgboost}
\begin{tabular}{lcccc}
\toprule
& \textbf{Precision} & \textbf{Recall} & \textbf{F1-score} & \textbf{Support} \\
\midrule
Non-Fraud [0] & 1.00 & 1.00 & 1.00 & 635441 \\
Fraud [1] & 0.98 & 0.70 & 0.82 & 821 \\
\midrule
Accuracy & & & 1.00 & 636262 \\
Macro avg & 0.99 & 0.85 & 0.91 & 636262 \\
Weighted avg & 1.00 & 1.00 & 1.00 & 636262 \\
\botrule
\end{tabular}
\footnotetext{\textbf{XGBoost model:} [[635427, 14], [245, 576]]}
\footnotetext{\textbf{Accuracy score on Test Data:} 0.9995929349859021\%}
\end{table}
\FloatBarrier

\subsection{Model Selection}\label{subsec:selection}

Random Forest was selected as the champion baseline model with the highest F1-score of 0.87.

\subsection{SMOTE Application}\label{subsec:smote}

To achieve a roughly equal class distribution in the training data, SMOTE was employed, increasing the dataset from 5.7 million to 11.4 million samples by creating synthetic examples of the minority class. This approach ensured adequate representation of the minority class during model training and mitigated bias towards the majority class.

\subsection{Hyperparameter Optimization Results}\label{subsec:hporesults}

The findings revealed that the enhanced Random Forest model achieved an accuracy of 99.97\%, a fraud precision of 92\%, a fraud recall of 90\%, and an F1-score of 0.91, significantly surpassing the baseline imbalanced model, demonstrating that the dual combination of SMOTE oversampling and hyperparameter optimization substantially enhanced fraud detection performance by addressing both data-level imbalance and algorithm-level bias.

\begin{table}[h]
\centering
\caption{Random Forest Model Performance}\label{tab:randomforest}
\begin{tabular}{lcccc}
\toprule
& \textbf{Precision} & \textbf{Recall} & \textbf{F1-score} & \textbf{Support} \\
\midrule
Non-Fraud [0] & 1.00 & 1.00 & 1.00 & 635441 \\
Fraud [1] & 0.92 & 0.90 & 0.91 & 821 \\
\midrule
Accuracy & & & 1.00 & 636262 \\
Macro avg & 0.96 & 0.95 & 0.95 & 636262 \\
Weighted avg & 1.00 & 1.00 & 1.00 & 636262 \\
\bottomrule
\end{tabular}
\footnotetext{\textbf{Random forest model:} [[635373, 68], [83, 738]]}
\footnotetext{\textbf{Accuracy score on Test Data:} 0.9997626763817421\%}
\end{table}
\FloatBarrier
\section{Conclusion and Future Work}\label{sec:conclusion}

This study illustrates that combining Random Forest with SMOTE and hyperparameter optimization is highly effective in identifying fraud within significantly imbalanced financial datasets, achieving an impressive 99.97\% accuracy and an F1-score of 0.91, which greatly exceeds the performance of baseline models. 

In operational FinTech settings, it is crucial that the model provides risk scores quickly to prevent delays in real-time transaction approvals. The Random Forest can be implemented as a scalable API microservice, capable of evaluating precomputed features such as type, amount, and balances in milliseconds, while managing peak demands through horizontal scaling. 

In adversarial situations such as fraud injection, where attackers test decision boundaries or alter patterns (such as dividing high-value transfers), the system needs ongoing drift monitoring, regular retraining, and hybrid protections that integrate machine learning with rule-based overrides to ensure robustness. 

Future research should explore temporal modeling (using LSTM/Transformers), multimodal features (like device and geolocation data), and causal inference to improve real-world performance. Additionally, hybrid ensembles that combine machine learning, rules, and domain expertise should be developed to balance fraud prevention with customer experience. Tools for explainability, such as SHAP or LIME, can further guarantee transparency, regulatory compliance, and expert validation of significant feature contributions.

\section{Acknowledgements}\label{sec:conclusion}
I am deeply grateful to Indira Gandhi Delhi Technical University for Women for offering a supportive and encouraging atmosphere. I also wish to extend my thanks to Dr. Ritesh Yaduwanshi, whose guidance, expertise, and unwavering support were instrumental in the success of this research. 

\backmatter

\bibliography{references}

\end{document}